# Time-Varying Formation Controllers for Unmanned Aerial Vehicles Using Deep Reinforcement Learning

Ronny Conde*. José Ramón Llata*. Carlos Torre-Ferrero*

*Electronics Technology, Systems and Automation Engineering Department at the University of Cantabria, 39005, Santander, Spain

(Corresponding Author: ronny.conde@gmail.com)

Abstract: We consider the problem of designing scalable and portable controllers for unmanned aerial vehicles (UAVs) to reach time-varying formations as quickly as possible. This brief confirms that deep reinforcement learning can be used in a multi-agent fashion to drive UAVs to reach any formation while taking into account optimality and portability. We use a deep neural network to estimate how good a state is, so the agent can choose actions accordingly. The system is tested with different non-high-dimensional sensory inputs without any change in the neural network architecture, algorithm or hyperparameters, just with additional training.

## 1. INTRODUCTION

Recently, formation control of unmanned aerial vehicles (UAVs) has attracted lots of researchers due to its broad potential applications, including search and rescue missions (Waharte et al., 2009) and wide area surveillance (Nigam et al., 2012). Considering these examples, it seems clear that widespread formation control systems will have a positive impact on society, from improving public security to saving lives after natural disasters.

For a UAV formation controller to become widespread the following requirements, among many others, should be met:

- Time-varying Formations: It should be able to work with time-varying formations, the most general case, to empower users to come up with new applications. Time-varying formations are those where relative separations and bearings change with time.

- Portability: It should generalize to any formation and work with different sensory inputs without requiring complex design processes. This requirement will help potential users deploy the system quickly and easily.

- Scalability: It should function properly with an arbitrarily large number of UAVs, which may be needed in some applications.

- Optimality: It should drive UAVs to reach formation as quickly as possible. Just guaranteeing stability may not be enough for some applications. Moreover, we want UAVs to do their best when desired formations are not feasible, because of UAV dynamic limitations, to help users design applications quickly, without checking formation feasibility.

Although many formation control strategies have been proposed, we have not found any approach that covers all these requirements.

A decentralized approach is needed for scalability. According to (Zhang and Mehrjerdi, 2013), it is less affected by computational and communicative bottlenecks. However, designing decentralized systems is difficult. We need global behaviors to emerge from individual ones without a central unit monitoring the accomplishment of the mission and sharing that information with team members.

It is so hard that most decentralized approaches study fixed formations. Only a few results have been obtained for time-varying formations, the most general case. These good results were achieved using consensus theory. Consensus theory can be used to analyze stability, getting necessary and sufficient conditions, but it is not evident, at least, how to guarantee optimality in the way we have defined it. Another issue of this approach is that complex design processes may be necessary when changing sensory inputs or the required formation.

Theoretically, a multi-agent system trained with classical reinforcement learning (RL) would meet most of the desired properties. It would be scalable because of the decentralized approach. Optimality requirement would be satisfied because RL is about agents learning optimal policies. However, it would be only partially portable. Even though the system could easily generalize to different time-varying formations, it would require complex design processes to come up with appropriate hand-crafted features when the sensory input changes.

In 2013, Mnih et al. presented a deep learning model able to learn control policies directly from sensory input, without relying on hand-crafted features, using reinforcement learning

(Mnih et al., 2013, Mnih et al., 2015). Now, thanks to this work, it seems clear that all desired requirements could be met by using this approach: *Deep Reinforcement Learning*.

1.1 Specific Limitations

For the sake of simplicity and to reduce computational needs, we do not consider obstacle avoidance, different communication topologies, delays, noisy sensors, complex UAV models or high-dimensional sensory inputs. All these elements will be part of future research.

1.2 Main Contributions

This paper supports that deep reinforcement learning is a feasible approach to develop scalable, optimal and portable time-varying formation controllers for UAVs.

- This brief confirms that deep reinforcement learning can be used to train individual UAVs to behave optimally in any time-varying formation, covering scalability and optimality.

- Even more, it supports that a deep reinforcement learning agent can be trained to drive UAVs to reach different time-varying formations in a plug-and-play fashion. This point partially covers portability.

- This work substantiates that generalizing to different sensory inputs can be done without complex design processes, just with additional training, at least for non-high-dimensional sensory inputs. Now, our definition of portability is covered.

- This paper demonstrates that this approach also works with unfeasible formations, because of UAV dynamic limitations.

2. PROBLEM DEFINITION

We developed a simulation environment to test the solution, including different formation sets and sensory inputs.

2.1 Simulation Environment

*Training / Test Mode:* During training, each simulated UAV receives a reward signal depending on how close it is to its desired position in the formation, according to the following equation:

$$r = -\sqrt{(x - x_g)^2 + (y - y_g)^2} \qquad (1)$$

Where $r$ is the reward signal, $(x, y)$ the position of the UAV and $(x_g, y_g)$ the desired position in the formation.

*UAV Model:* The simulator models rotary-wing UAVs, like quadrotors or vertical take-off and landing aircrafts (VTOL). Since the trajectory dynamics has much larger time constants than the attitude dynamics, the formation control can be implemented with an inner/outer loop structure. The inner loop controls the attitude, and the outer loop is used to drive the UAVs to the desired positions (Dong et al., 2015, Karimoddini et al., 2013).

Because our agent controls the outer-loop, each UAV can be modeled as a point-mass system described by the following double integrator (Dong et al., 2015):

$$\begin{cases} \dot{x}(t) = v_x(t) \\ \dot{y}(t) = v_y(t) \\ \dot{v}_x(t) = u_x(t) \\ \dot{v}_y(t) = u_y(t) \end{cases} \qquad (2)$$

Where $(x, y)$ denotes the position of the UAV, $(v_x, v_y)$ represents its velocity and $(u_x, u_y)$ the control inputs.

*Two-Dimensional Environment:* For the sake of simplicity, all quadrotors move in the XY plane.

*Labelled Homogeneous Robots:* All UAVs are homogeneous but not interchangeable. Each UAV knows its desired position in the formation at every time step.

*Discrete Control Commands:* The controller chooses actions from a discrete set $\mathcal{A} = \{\rightarrow, \leftarrow, \uparrow, \downarrow, \square\}$ at each time step, where "→" represents positive acceleration on the x axis, "←" represents negative acceleration on the x axis, "↑" represents positive acceleration on the y axis, "↓" represents negative acceleration on the y axis and "□" represents no acceleration.

*Formation Training Set, $\mathcal{F}_{training}$:* The simulation environment provides different time-varying formations for the training of our controller.

*Formation Test Set, $\mathcal{F}_{test}$:* The formations included in this set are different from the ones included in $\mathcal{F}_{training}$ because we want to test whether the approach generalizes to any time-varying formation. We have also included non-feasible formations to prove that the system can handle them.

*Sensory Inputs:* Two sensory inputs are provided to check if the solution satisfies the portability requirement:

- Precise localization system. Each UAV knows its position and the velocity estimation at each time step.

- Distance to four landmarks. Each UAV knows the distance to four landmarks and the velocity estimation at each time step.

For the sake of simplicity, and to reduce computational needs, we work with non-high dimensional sensory inputs. High-dimensional ones would require a model with higher capacity.

*Episodic Environment:* Every four hundred time steps the simulation is reset, placing each simulated UAV in a new start position and choosing a new formation from the appropriate set.

2.2 Checking Requirements

The following checklist has been used to verify the requirements:

- Optimality. The controller should try to maximize cumulative reward on each episode, reaching formation as quickly as possible and keeping it afterward.

- Portability. The controller should be trained with formations from $\mathcal{F}_{training}$ and work properly with formations from $\mathcal{F}_{test}$ if sensory input remains the same. Even more, the system should work in the same way with another sensory input just with additional training.
- Scalability. A different controller instance should be loaded on each UAV. No central unit is allowed.
- Time-varying and non-feasible formations were included in $\mathcal{F}_{test}$.

# 3. DEEP REINFORCEMENT LEARNING AS A FEASIBLE APPROACH

RL is the area of machine learning concerned with how software agents ought to take actions based on experience. During training, the agent receives a reward signal at each time step, used to define its goal, and learns how to maximize the total reward it receives.

## 3.1 Training Cycle

During training, the following cycle is repeated until the episode ends:

- The agent combines the UAV sensory input with the information about its role in the formation into a state $s$. State is defined as the information taken into account by the agent to choose the next action.
- The agent picks action $a$, from the available actions set $\mathcal{A}$, given the current state, $s$.
- After executing the selected action, the agent ends up in a new state, $s'$, and receives the appropriate reward, $r$. $s'$ was obtained by combining the new sensory data with the formation specification.

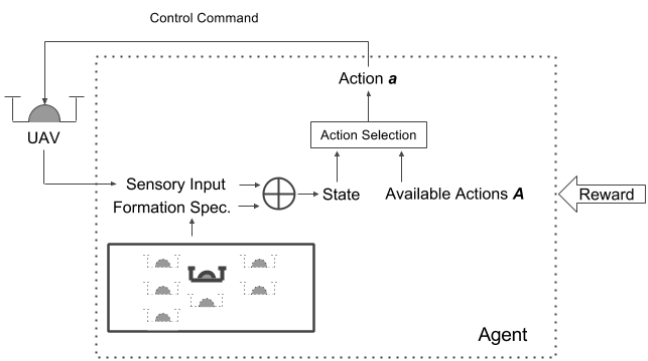

Fig 1. Training Cycle, repeated until the episode ends.

## 3.2 Discount Factor and Stochasticity

RL takes into account stochasticity. Ending up in a new state $s'$ is partly random and given by the transition function $T(s, a, s') = P(s' | s, a)$, the probability of ending up in $s'$ given the prior state and chosen action. As a result, we may not receive the same rewards by choosing the same actions. For this reason, we use discounted rewards to calculate the expected total reward until the end of the episode:

$$R_t = \sum_{i=t}^{n} \gamma^{(i-t)} r_i \qquad (3)$$

where $t$ is the current time step, $n$ is the length of the episode, and $\gamma$ is a real number between 0 and 1, called discount factor, that reflects how much the agent takes into account future rewards.

## 3.3 Q-values and Optimal Behavior

A Q-state represents the commitment of executing an action from a state. E.g. $\{(s, \leftarrow), (s, \uparrow), (s, \rightarrow), (s, \downarrow), (s, \square)\}$ is the list of the Q-states associated to a hypothetical state $s$. A value is assigned to each state-action pair, Q-value or $Q(s, a)$, to estimate how good is choosing an action from a state.

$Q(s, a)$ maps each Q-state to the expected total discounted reward if the agent executes action $a$ from state $s$ and behaves optimally afterward. E.g. $Q(s, \leftarrow)$ returns the expected total discounted reward, $\mathbb{E}[r_t + \gamma r_{t+1} + \gamma^2 r_{t+2} + \ldots]$, after choosing action "$\leftarrow$" from state $s$ and behaving optimally afterward, selecting the actions with the highest Q-value.

Once all Q-values are calculated, the optimal behavior can be easily extracted. The agent just needs to choose the action with the highest associated Q-value at each time step to be optimal.

## 3.4 Deep Reinforcement Learning

The problem implies a continuous state space but, how can an infinite amount of Q-states be stored? A lookup table cannot be used. Besides, the agent needs to be able to estimate Q-values of state-action pairs it has not visited before.

Classical approaches use linear combinations of hand-crafted features to approximate $Q(s, a)$. However, coming up with good hand-crafted features is a tough problem because $Q(s, a)$ usually has strong non-linearities. This approach does not meet our portability requirement because changes in the sensory input would require a complex design process to come up with new hand-crafted features.

In this work, a different approach to estimating Q-values is used: deep neural networks. Deep neural networks can learn useful features by themselves if enough data is provided. Even more, according to (Krizhevsky et al., 2012), they usually learn better representations than hand-crafted features. As a result, the agent can work with different sensory inputs just with additional training.

## 3.5 Epsilon-Greedy

Before diving into how to train the neural network, the exploratory strategy during training needs to be considered.

If the agent always chooses the greedy action, the one with the highest Q-value, it could get stuck in suboptimal strategies.

To avoid this problem, epsilon-greedy is used as the exploratory strategy. A random action is chosen with a small probability $\varepsilon$, picking the greedy one otherwise.

## 3.6 Deep Neural Network Training

Since $Q(s,a)$ are real values, the neural network can be trained like in a regression problem.

The Bellman optimality equation is used to come up with an appropriate loss function:

$$Q(s,a) = \mathbb{E}_{s'}[r + \gamma Max_{a' \in \mathcal{A}} Q(s',a')|s,a] \quad (4)$$

The Loss function was calculated by using (4):

$$L = \frac{1}{2}[r + \gamma Max_{a' \in \mathcal{A}} Q(s',a') - Q(s,a)] \quad (5)$$

However, applying it naively has several problems, including:

- Learning from consecutive samples is inefficient, due to the strong correlations between them.
- Since learning the current parameters determines the next data sample, the process could be unstable.

To solve these problems, we use a technique called Experience Replay (Lin, 1993). The last N agent´s transitions are stored in a dataset called Replay Memory and, at each time step, samples of transitions are drawn at random from the Replay Memory to train the network. This solution stabilizes the whole training process.

## 3.7 How the Approach Solves the Problem

Scalability. A different agent instance is in charge of each UAV.

Optimality. RL is about software agents learning how to behave optimally. Optimality is defined by a reward strategy, and the agent learns how to behave optimally with training.

Portability. Thanks to the rich state space our agent generalizes to any time-varying formation without any additional process. Because of the use of a deep neural network to approximate Q-values, our agent works with different sensory inputs just with additional training.

## 4. SYSTEM SPECIFICATION

The same neural network architecture and hyperparameters were used across all scenarios, needed for portability.

### 4.1 Neural Network Architecture

In addition to the input layer, which depends on the state representation, the model has three fully-connected hidden layers, with 128, 64 and 32 rectified linear units (ReLU) respectively. The output layer is a fully-connected one with a single output, the approximation of $Q(s,a)$.

We used Keras (Chollet, 2015) on top of Theano (Team et al., 2016) to build and train the deep neural network. We selected these platforms because of Keras' focus on fast experimentation, Theano's ability to run on GPU and the availability of a Python API, our development language.

We chose RMSProp as the optimizer and "Uniform" as the weight initialization strategy. Specific details are presented below.

### 4.2 Hyperparameters

*Epsilon-greedy module:* $\varepsilon$ was equal to 0.5 during training.

*Replay Memory Size:* It stores the $10^6$ most recent transitions.

*Replay Memory Start Size:* The agent followed a random policy for $10^4$ transitions to populate the replay memory before learning starts.

*RMSProp Parameters:* $5 \cdot 10^{-6}$ was used as a learning rate, $10^{-8}$ as epsilon (Small value added for numerical stability) and 0.9 as rho (Gradient moving average decay factor).

*Discount Factor:* $\gamma$ was set to 0.95.

*Minibatch Size:* We used 16 training samples to compute each SGD update.

### 4.3. Reward Clipping and Normalization

The reward function (1) offered by the simulation environment lacks a lower limit. Two normalization steps have been applied to avoid large errors:

1. The reward has been increased by 1. After this operation, the new upper limit is 1.
2. The reward has been clipped to [-1, +1]

## 5. MAIN EVALUATION

### 5.1 Generalization to Different Non-High-Dimensional Sensory Inputs

The agent was trained using both sensory inputs provided by the simulation environment: precise localization system and distance to three landmarks. The total clipped reward collected per episode was used as the primary metric to evaluate the progress of the agent. This metric tends to be very noisy, especially when $\varepsilon$ is set to a relatively high value, 0.5. A Savitzky-Golay filter has been used in (Fig. 2) and (Fig 3.) to smooth the data and mitigate this problem.

The upper limit of the metric in (Fig. 2) and (Fig. 3) is 400 because the maximum normalized reward is 1.0 per time step and the simulation environment resets every four hundred time steps. However, the following points make it impossible to reach the upper limit during training:

- $\varepsilon$ has been set to 0.5, a relatively high value. It means that, during training, a random action, instead of the greedy one, is chosen with a probability of 0.5 at every time step.
- The actions are chosen from a discrete set $\mathcal{A} = \{\rightarrow, \leftarrow, \uparrow, \downarrow, \square\}$, instead of using a continuous action space.
- A small neural network has been used to avoid complexity and computational needs, so the capacity of the model is relatively small.

However, despite all these limitations, (Fig 2.) and (Fig. 3) show steady learning processes in both scenarios.

The same neural network, architecture, hyperparameters and algorithm have been used in both situations, including the learning rate and the epsilon-greedy values. (Fig. 2) and (Fig. 3) display similar results, with the exception that the second scenario, distance to three landmarks, requires more training to get to similar cumulative rewards, 1500 episodes instead of 1000.

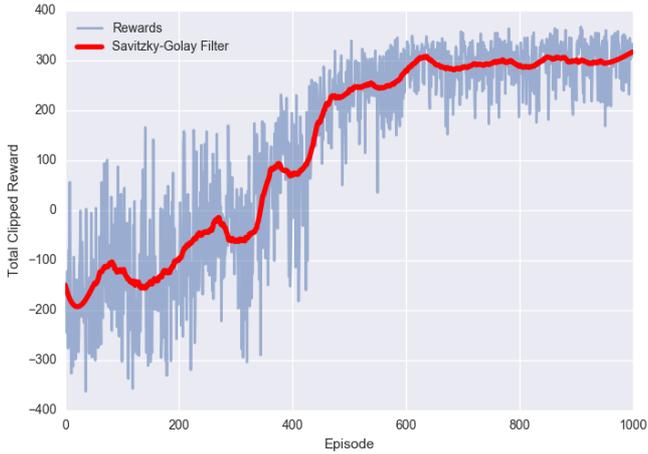

Fig. 2. Total clipped reward collected per episode using a precise localization system as sensory input.

These results support the idea that deep reinforcement learning can be used to build agents able to generalize to different sensory inputs with just additional training.

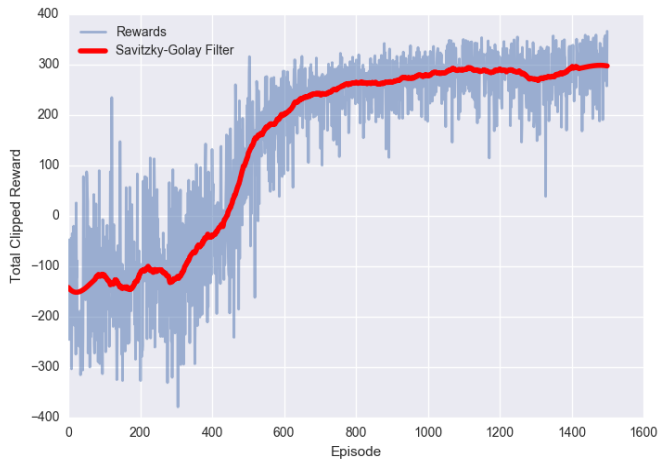

Fig. 3. Total reward collected per episode using the distance to four landmarks as sensory input.

### 5.2. Generalization to Different Time-Varying Formations

The agent was trained with formations drawn at random from $\mathcal{F}_{training}$. In this section we want to find out how the agent behaves with formations from $\mathcal{F}_{test}$, formations it has not seen before.

To be more specific, (Fig. 4) shows the trajectories of five UAVs following an eight-figure pattern while keeping a phase separation of $2\pi/5$ for 400 time steps, the episode length. This formation is similar to the one used in (Dong et al., 2015). Each simulated UAV was controlled by an instance of the trained agent.

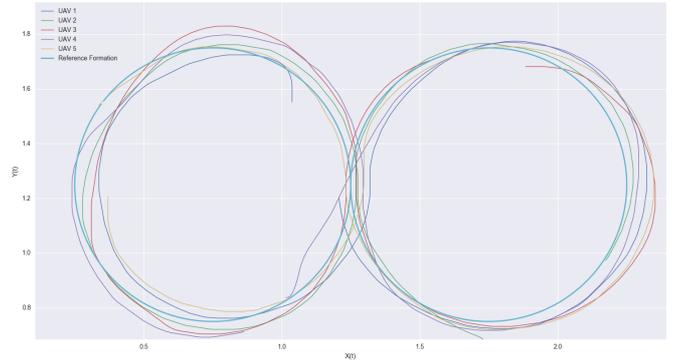

Fig. 4. Trajectories of five UAVs trying to follow an eight-figure pattern while keeping a phase separation of $2\pi/5$.

Despite the fact that all formations included in $\mathcal{F}_{test}$ are not feasible because of the discrete action space, (Fig. 5) shows how the system handles a more obvious non-feasible formation. It would be unfeasible even with a continuous action space because of the acute angles of the pattern. Again, each simulated UAV was controlled by an instance of the trained agent.

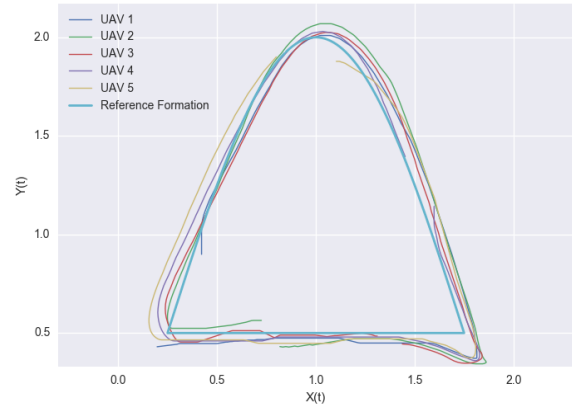

Fig. 5. Trajectories of five UAVs trying to reach a non-feasible time-varying formation while keeping a phase separation of $2\pi/5$.

## 6. RELATED WORK

The related work is analyzed from two different perspectives: UAV formation control and Deep Reinforcement Learning.

### 6.1 UAV Formation Control

*Centralized vs. Decentralized Architecture*: The decentralized approach was highly influenced by (Zhang and Mehrjerdi, 2013). This survey reported that centralized systems do not scale well as formation size increases because of communication bottlenecks and the lack of use of the computational resources available on each vehicle. This was stated to be true even when the most advanced optimization solvers are used.

*Time-Varying Formations:* Most decentralized approaches to UAV formation control study fixed formations. Few successful approaches have been reported for time-varying formations, the most general case. These good results were achieved using consensus theory, like (Dong et al., 2015) or (Rui et al., 2015). Consensus theory can be used to analyze stability but is not evident, at least, how it can be used to guarantee optimality in the way we have defined it. Another problem with this approach is that complex design processes may be necessary when changing sensory inputs or the required formation.

6.2 Deep Reinforcement Learning

Combining RL algorithms with nonlinear function approximators, like neural networks, could cause the training process to diverge, so the vast amount of work in RL used to focus on linear function approximators (Tsitsiklis and Van Roy, 1997).

The main problem when using RL with neural networks is that RL agents incrementally update their parameters while they observe a stream of experience, breaking the independent and identically distributed assumption of many stochastic gradient-based algorithms.

However, in 2013, Mnih et al. presented a deep learning model able to learn control policies directly from sensory input (Mnih et al., 2013, Mnih at al., 2015). They did it using experience replay (Lin, 1993), which addresses the stability problem by mixing more and less recent experience when updating the neural network weights.

Thanks to this work, it seems clear that this approach can be used to meet portability because hand-crafted features are not needed anymore. Even more, according to (Krizhevsky et al., 2012), deep neural networks can, usually, learn better representations than hand-crafted features if enough data is provided.

## 7. CONCLUSIONS

This paper confirmed the feasibility of using deep reinforcement learning to develop scalable, optimal and portable time-varying formation controllers for UAVs.

Scalability requirement was met because an agent was trained to control individual UAVs and different instances were installed on each vehicle.

Optimality was also considered. The reward strategy was used to define what we meant by optimal, and the agent learned how to behave optimally with training.

Besides, we showed that our system generalizes to different time-varying formations and confirmed that is able to work with different non-high-dimensional sensory inputs just with additional training, thanks to a deep neural network to approximate $Q(s,a)$.

We did not consider important aspects of UAV formation controllers, including obstacle avoidance, complex UAV models and high-dimensional sensory inputs. These topics will be part of future research.